\pdfoutput=1
\documentclass{llncs}
\usepackage[T1]{fontenc}
\usepackage{graphicx}
\usepackage{booktabs}
\usepackage[table]{xcolor}
\usepackage{subcaption}
\usepackage{threeparttable}
\usepackage{multirow}
\usepackage{xspace}
\usepackage{amsmath}
\usepackage{enumitem}
\usepackage{url}

\newcommand{\this}{\textsc{VeriScene}\xspace}
\newcommand{\revise}[1]{#1}

\begin{document}
\pagestyle{empty}

\title{\this: Reconstructing Crime Scenes from Legal Evidence via World-Model Agent}
\titlerunning{Reconstructing Crime Scenes from Legal Evidence}
\author{Kevin Chuanpu Fu\inst{1} \and
Yongsen Zheng\inst{1} \and
Zee Kin Yeong\inst{2} \and
Kwok-Yan Lam\inst{1}\thanks{Corresponding author.}}
\authorrunning{K. C. Fu et al.}
\institute{Nanyang Technological University, Singapore \and
Singapore Academy of Law, Singapore}

\maketitle
\thispagestyle{empty}

\begin{abstract}
World models take multimodal inputs like text, photos, and diagrams to generate dynamic scenes in accordance with the laws of physics, thus opening a compelling application: fusing multimodal legal evidence to re-create a crime scene and re-enact how an offence could have been committed.
However, feeding the raw, unorganized evidence into a world model fails in forensic use: it silently drops evidence, glosses over contradictory testimony, and produces motion that violates the evidentiary record.
This paper presents \this, an agent that orchestrates the world model: it reconstructs crime scenes from forensic photographs and witness statements of varying reliability, keeping every claim traceable to evidence and every motion physically plausible.
\this iteratively fuses the evidence into a cited narrative under an auditing loop, verifies the hypothesized dynamics via probe rollouts in the world model with corrective constraint injection, and renders the offence as a re-enactment video from a fused keyframe.
On a benchmark of 25 crime scenarios across 7 physically-driven case types (139 forensic-style photographs and 65 statements with planted unreliability), \this attains 0.9014 evidence coverage and 0.7217 factual consistency (0--1 scale) on the 20 test scenes, outperforming an end-to-end multimodal-LLM baseline by 20.35\% in factual consistency and 34.88\% in temporal coherence, while generalizing across four LLM orchestration backends at \$1.82 per scene.

\end{abstract}

\section{Introduction} \label{section:introduction}

Adjudication rests on evidence of many kinds: scene photographs, seized-object records, identification photos, and witness statements of varying reliability.
To weigh them, judges and jurors must mentally re-create how the alleged offence unfolded---yet such reconstruction is subjective conjecture: statements conflict, evidence scatters across modalities, and a mental picture obeys no objective physical law, while misreading a single physical cue (e.g., a glass-fracture direction) can invert an entire case.
World models~\cite{Web-Veo,Web-Cosmos} take multimodal inputs and generate dynamic scenes governed by physics, making them a natural instrument for replacing conjecture with a watchable, physically grounded re-enactment.

In this paper, we propose an agent that orchestrates the world model for crime-scene reconstruction: it automatically fuses the heterogeneous evidence into a coherent, evidence-cited account, and iteratively reasons over the draft in an audit loop that eliminates ambiguity and resolves contradictions before any frame is rendered.

Directly prompting generative models with case material fails here: evidence enters without order or organization, so relations among exhibits---which statement corroborates which photograph, which trace arbitrates which conflict---are never modeled; the output drops evidence, ignores contradictions~\cite{NeurIPS23-LLaVA}, and favors visual appeal over physical correctness~\cite{ICLR25-VideoPhy}, leaving no artifact by which errors can be audited back to their inducing evidence.

Inspired by how investigators draft a hypothesis, cross-examine it against the evidence, and test its physical feasibility, we posit that reconstruction quality is governed by iterative verification rather than model scale: citation audits catch uncovered evidence, world-model probes catch physically invalid dynamics, and verified-physics injection catches unfaithful rendering.

We therefore introduce \this, which turns multimodal legal evidence into physically faithful crime-scene re-enactments: it fuses the evidence into a grounded narrative with per-claim citations, distills a physics-annotated motion specification, and renders the re-enactment video from a fused keyframe---each stage guarded by its own verification loop.

Realizing this pipeline raises one challenge per module: \emph{comprehensiveness}, met by an auditing agent issuing targeted follow-ups until every evidence item and contradiction is covered; \emph{physical faithfulness}, met by probing the hypothesized key event in the world model and injecting corrective constraints; and \emph{faithful rendering}, met by compiling the verified constraints into the generation prompt with completeness and anti-exaggeration clamps.

To assess \this, we construct a benchmark of 25 crime scenarios across 7 physically-driven case types (139 forensic-style photographs, 65 witness statements with planted unreliability).
\this achieves 0.9014 evidence coverage and 0.7217 factual consistency (0--1 scale) on 20 test scenarios, outperforming an end-to-end multimodal-LLM baseline by 20.35\% in factual consistency and 34.88\% in temporal coherence, and generalizing across five LLM backends at \$1.82 per scene.

In summary, the contributions of this paper are three-fold:
\begin{itemize}[nolistsep]
    \item \textbf{New problem.} We formulate crime-scene reconstruction as orchestrating a world model to fuse multimodal legal evidence under physical and evidentiary constraints.
    \item \textbf{New system.} We design \this, an agent pipeline of iterative evidence fusion, world-model physical reasoning, and physics-annotated re-enactment rendering, each with its own verification loop.
    \item \textbf{New benchmark.} We construct a 25-scenario, 7-case-type multimodal legal evidence benchmark with five decimal-valued metrics\footnote{Code and benchmark: \url{https://github.com/fuchuanpu/VeriScene}.}, and validate \this against baselines, ablations, and five LLM backends.
\end{itemize}

\section{Problem Statement} \label{section:statement}

\noindent\textbf{Task Setting.}
An investigator (analyst) must reconstruct the course of an alleged offence from an evidence set $E$ collected at the scene: (i) forensic photographs depicting the post-incident state and (ii) natural-language witness statements from individual viewpoints.
However, $E$ is not a consistent record: photographs capture only static end states, at least one of the two to three statements per scene is partially unreliable and may contradict other evidence, and no ground-truth footage exists, so the reconstruction must be justified by the evidence itself.

Given $E$, the analyst must produce an evidence-grounded narrative---every asserted event traceable to supporting photographs or statement fragments---and a re-enactment video rendering it as physically plausible motion; when the reconstruction rests on an unresolved contradiction, \this reports the conflicting items rather than silently committing to one account.

\noindent\textbf{Design Goals.}
\this should meet three requirements:
(1)~\textit{Comprehensiveness.}
Every item in $E$ enters the reconstruction, since omitted evidence invalidates a legal account.
(2)~\textit{Physical faithfulness.}
The reconstructed dynamics and rendering should obey physical constraints (e.g., momentum transfer, gravity) that generative video models frequently violate~\cite{ICLR25-VideoPhy}.
(3)~\textit{Verifiability.}
Every claim and rendered event should be attributable to specific evidence items, with contradictions and reliability judgments reported explicitly for human audit.

\section{System Overview} \label{section:overview}

\begin{figure}[!t]
    \centering
    \includegraphics[width=0.93\textwidth]{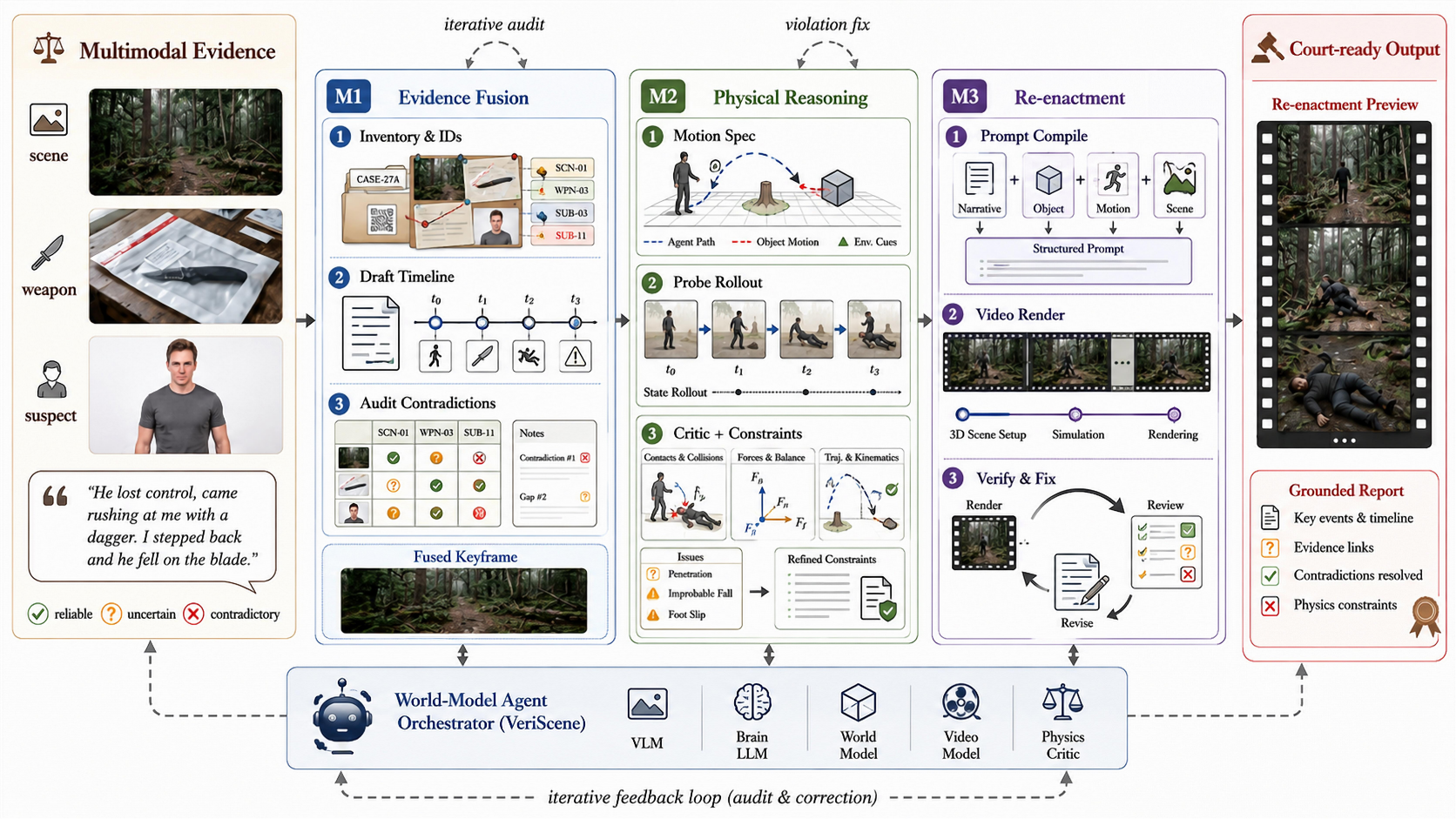}
    \caption{Architecture of \this.}
    \label{diagram:framework}
    \vspace{-4mm}
\end{figure}

Handing the entire evidence set to a multimodal model in one prompt fails on three counts: details appearing in only one photograph are silently omitted, contradicting statements are blended into a fluent narrative instead of weighed against the physical evidence, and text-conditioned generation is not physically grounded~\cite{ICLR25-VideoPhy}.
Yet each failure is \emph{checkable}---dropped evidence by auditing against an evidence inventory, contradictions by cross-examination, and physical violations by probing candidate dynamics in a world model and critiquing the rollout~\cite{NeurIPS23-Reflexion}---\revise{\emph{world model} here follows the learned-simulator sense of Ha--Schmidhuber~\cite{NeurIPS18-WorldModels} and LeCun~\cite{Web-LeCunWM}, in its video-generative branch~\cite{Web-Cosmos}}.
\this therefore decomposes reconstruction into one generation--verification loop per failure mode, chained as the three modules in Figure~\ref{diagram:framework}.

\noindent\textbf{Iterative Evidence Fusion.}
\this builds an evidence inventory with an identifier per photograph and statement fragment, a VLM fuses it into a draft scene description, and an auditor agent flags uncovered items, uncited claims, and contradictions, driving bounded revisions until every claim carries supporting identifiers.

\noindent\textbf{World-Model Physical Reasoning.}
This module tests the fused description's candidate dynamics physically: each candidate is rendered as a short probe rollout in the world foundation model~\cite{Web-Cosmos}, a physics critic inspects it, and detected violations become constraints injected into the next probe, terminating with validated constraints attached to the narrative.

\noindent\textbf{Re-enactment Rendering.}
Finally, \this compiles the audited narrative and validated constraints into a physics-annotated prompt rendered by the video world model~\cite{Web-Veo}; inheriting identifiers and constraints, every rendered event remains traceable to its supporting evidence.

\section{Design of \this} \label{section:design}

\subsection{Benchmark Construction} \label{section:design:dataset}

\noindent\textbf{Scenario Schema and Generation.}
As no public dataset couples multimodal legal evidence with a verifiable physical ground truth, we build 25 synthetic scenarios spanning seven case types (Table~\ref{table:dataset}): 20 held-out evaluation scenes (111 photographs, 53 statements, 33 unreliable) plus a five-scene development split.
A frontier LLM (Claude~\cite{Web-Claude3}) generates each scenario, under the guidance of legal practitioners on evidence types and investigative conventions, against a programmatically validated schema separating the pipeline-visible \emph{evidence bundle} from a hidden \emph{ground truth} (factual summary, 4--6-event timeline, and the \emph{key physical fact} resolving the case); every case is physical-mechanism-driven with 1--2 distractors.

\begin{table}[!t]
\centering
\caption{Statistics of the 25 constructed crime scenarios.}
\label{table:dataset}
\scriptsize
\setlength\tabcolsep{1.1mm}
\begin{tabular}{lcccccc}
\toprule
Case type & Scenes & Images & Person/Obj./Scene & Statements & Unreliable & Avg.\ words \\
\midrule
Physical altercation & 3 & 17 & 5/6/6 & 6 & 3 & 189.0 \\
\rowcolor{gray!12} Arson origin & 3 & 16 & 3/6/7 & 9 & 6 & 176.8 \\
Burglary path & 4 & 23 & 4/10/9 & 8 & 4 & 179.5 \\
\rowcolor{gray!12} Water incident & 3 & 16 & 4/6/6 & 8 & 5 & 175.1 \\
Fall from height & 4 & 21 & 5/8/8 & 11 & 7 & 197.4 \\
\rowcolor{gray!12} Property damage & 4 & 21 & 4/8/9 & 11 & 7 & 178.0 \\
Traffic incident & 4 & 25 & 4/11/10 & 12 & 8 & 172.8 \\
\midrule
Total & 25 & 139 & 29/55/55 & 65 & 40 & 181.0 \\
\bottomrule
\end{tabular}
\vspace{-4mm}
\end{table}

\noindent\textbf{Evidence Image Synthesis.}
Each scenario specifies 4--7 photographs of four forensic types---\emph{victim}/\emph{perpetrator} identification shots, \emph{object} close-ups beside a ruler scale, and 2--3 wide \emph{scene} shots with numbered markers---rendered by an image model~\cite{Web-Gemini} under a shared template enforcing sober documentation style and a non-graphic safety suffix.
Mirroring forensic practice, persons are photographed \emph{after} rescue, so every image is individually innocuous while the incident remains recoverable from physical traces (e.g., skid marks).

\noindent\textbf{Planted Testimony Unreliability.}
Each scenario carries 2--3 first-person statements (150--250 words, police-record register): one reliable, one with a planted time or position discrepancy, and optionally one contradicting another witness (40 of 65 overall); recorded discrepancies and resolutions test whether conflicts are settled by physical evidence rather than averaged testimony.

\subsection{Multimodal Evidence Fusion} \label{section:design:fusion}

\noindent\textbf{Structured Fusion Drafting.}
The first module distills the evidence bundle---never the ground truth---into a structured reconstruction: a frontier VLM (Gemini~\cite{Web-Gemini}) receives all photographs, captions, and statements under an arbitration rule that photographic and physical evidence overrides testimony \revise{(mirroring investigative practice, which treats photographs with greater objectivity than statements)}, and returns a \emph{narrative} with per-claim evidence traceability, a \emph{timeline} of 4--6 steps citing supporting files, surfaced \emph{contradictions} with physical resolutions, and a \emph{keyframe prompt} plus a one-sentence \emph{motion hint} covering the key event through its outcome.

\noindent\textbf{Audit-and-Refine Loop.}
The draft is audited by the LLM (Claude), a directed critic in the spirit of self-refinement~\cite{NeurIPS23-Reflexion} checking that every file is cited or explicitly irrelevant, every contradiction is physically resolved, and the timeline is physically ordered; it returns at most three \emph{directed follow-ups}, re-injected append-only, for at most three drafting rounds with early stopping.

\noindent\textbf{Documentation-Protocol Grounding.}
Because victims are photographed unharmed by protocol, an unprimed fusion model repeatedly read their intact appearance as \emph{exculpatory}; the prompt therefore declares the acquisition protocol explicitly---otherwise its meaning inverts.

\noindent\textbf{Keyframe Synthesis.}
The keyframe is rendered by the image model conditioned on up to three original scene photographs, inheriting the documented environment; an open text-to-image fallback~\cite{Web-Cosmos} covers refusals.

\subsection{World-Model Physical Reasoning} \label{section:design:reasoning}

\begin{figure}[!t]
    \centering
    \includegraphics[width=0.99\textwidth]{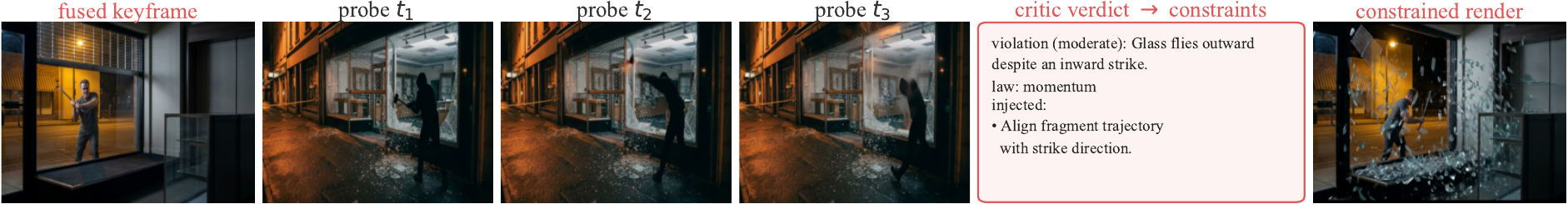}
    \caption{Probe-and-correct physical verification on a real scene.}
    \label{graph:probemech}
    \vspace{-5mm}
\end{figure}

\noindent\textbf{Physics Motion Specification.}
The second module validates intended motion \emph{before} expensive rendering: the LLM, prompted as a physical-dynamics expert, maps the narrative, timeline, and motion hint to an 8-second image-to-video specification---a \emph{video prompt} (who moves, where, at what speed, cause before effect), 3--5 \emph{physics constraints} (e.g., free-fall acceleration), and 3--5 \emph{negative terms} naming failure modes to suppress (e.g., teleporting, floating).

\noindent\textbf{Probe-and-Correct Verification.}
Rather than critiquing text, \this critiques \emph{motion} (Figure~\ref{graph:probemech}): an open video world model (Cosmos~\cite{Web-Cosmos}) renders a low-resolution probe clip ($832{\times}480$, ${\sim}2$\,s) from the keyframe, and the VLM critic---scoped to \emph{physics-law violations only}~\cite{ICLR25-VideoPhy} (e.g., glass shards flying outward against an inward strike)---turns its severity-ranked verdict into at most two added constraints and two negative terms, merged append-only.
The probe is best-effort, falling back to the unrefined specification, never blocking the pipeline.

\subsection{Scene Re-enactment Rendering} \label{section:design:render}

\noindent\textbf{Physics-Annotated Prompt Assembly.}
The rendering prompt concatenates the video prompt, the probe-refined constraints, and three fixed clamps: \emph{completeness} (the full event through its physical outcome), \emph{anti-cinematic} (motion follows the stated physics exactly under fixed-camera documentary framing), and \emph{non-graphic} (restating the safety constraints of Section~\ref{section:design:dataset}).

\noindent\textbf{Primary Rendering with Safety-Aware Retry.}
A frontier video world model (Veo~3.1~\cite{Web-Veo}) generates the 8-second clip image-to-video from the keyframe, with per-model quota failover; when incident semantics trip the hosted safety filter, \this retries once with the content reframed as a fictional re-enactment staged by professional actors.

\noindent\textbf{Open-Model Fallback and Frame Extraction.}
If both attempts fail, rendering falls back to the open world model~\cite{Web-Cosmos} at probe resolution but full length, reusing the specification and blacklist; eight uniformly spaced frames are then extracted for the judge of Section~\ref{section:evaluation} and the qualitative storyboard.

\section{Evaluation} \label{section:evaluation}
We evaluate \this on crime scenarios spanning seven case types, analyzing reconstruction quality against baselines and ablations (Section~\ref{section:evaluation:overall}), qualitative fidelity (Section~\ref{section:evaluation:qualitative}), and the impact of the orchestration LLM (Section~\ref{section:evaluation:llm}).

\subsection{Experiment Setup} \label{section:evaluation:setup}
\noindent\textbf{Implementation.}
We prototype \this in Python on the claude-agent-sdk: evidence fusion and self-critique use Gemini (\texttt{gemini-3.1-pro}), keyframes are synthesized by \texttt{gemini-3-pro-image}, and video segments are rendered by Veo~3.1 on its fast tier; when the hosted renderer is unavailable, a Cosmos3-Nano low-resolution fallback runs locally on one NVIDIA H200 GPU; the weak orchestration tier (Section~\ref{section:evaluation:llm}) runs entirely on this local stack (Cosmos text-to-image keyframes and image-to-video renders), all other tiers as above.
The judge is Gemini, decoupled from the orchestration brain to avoid self-preference.
All reported runs cover the 20 held-out main scenes (the 5 development scenarios are used only for prompt tuning), and the orchestration-tier study (Section~\ref{section:evaluation:llm}) uses the first 10, except its weak tier.

\noindent\textbf{Metrics.}
We score each reconstruction with five metrics on a 0--1 scale: Evidence Coverage Rate (ECR), the fraction of evidence files explicitly cited in the fused timeline (computed programmatically); Factual Consistency (FC), the fraction of ground-truth timeline events reflected in the narrative, capped at 0.5 if the key physical fact is missed; Temporal Coherence (TC), the fraction of adjacent ground-truth event pairs shown in correct order across 8 frames extracted from the rendered video; and Physical Plausibility (PP) and Visual Fidelity (VF), scored 1--5 by the LLM-as-judge protocol~\cite{NeurIPS23-MTBench} and normalized to 0--1.

\noindent\textbf{Baselines.}
We compare against five naive world-model baselines---\textit{Text-to-video} (concatenated captions and statements prompted directly to the local world model), \textit{Raw-prompt I2V} (the same text with the first scene photo as the starting frame), \textit{LLM-prompt I2V} (one cheap text-only LLM call writes the video prompt), and the single-modality \textit{Caption-only I2V} and \textit{Testimony-only T2V}---as well as \textit{E2E-LLM}, which feeds all evidence into a single multimodal prompt and renders with the same video model; \textit{w/o Fusion}, which replaces the iterative fusion audit with single-shot fusion; and \textit{w/o Reasoning}, which removes the world-model probe, so no physical-consistency constraints are injected before rendering.

\subsection{Overall Performance} \label{section:evaluation:overall}
\begin{table}[!t]
\centering
\caption{Reconstruction quality of \this and baselines.}
\label{table:main}
\scriptsize
\setlength\tabcolsep{1.2mm}
\begin{tabular}{lccccc}
\toprule
Method & ECR & FC & PP & TC & VF \\
\midrule
Text-to-video & 0{\scriptsize$\pm$0.00} & 0.5083{\scriptsize$\pm$0.26} & 0.45{\scriptsize$\pm$0.11} & 0.02{\scriptsize$\pm$0.09} & 0.48{\scriptsize$\pm$0.10} \\
\rowcolor{gray!12} Testimony-only T2V & 0{\scriptsize$\pm$0.00} & 0.5483{\scriptsize$\pm$0.29} & 0.44{\scriptsize$\pm$0.10} & 0.09{\scriptsize$\pm$0.25} & 0.49{\scriptsize$\pm$0.12} \\
Raw-prompt I2V & 0{\scriptsize$\pm$0.00} & 0.5167{\scriptsize$\pm$0.28} & 0.44{\scriptsize$\pm$0.10} & 0.01{\scriptsize$\pm$0.04} & 0.54{\scriptsize$\pm$0.11} \\
\rowcolor{gray!12} Caption-only I2V & 0{\scriptsize$\pm$0.00} & 0.2983{\scriptsize$\pm$0.21} & \textbf{0.55{\scriptsize$\pm$0.15}} & 0{\scriptsize$\pm$0.00} & 0.58{\scriptsize$\pm$0.15} \\
LLM-prompt I2V & 0{\scriptsize$\pm$0.00} & 0.6417{\scriptsize$\pm$0.19} & 0.41{\scriptsize$\pm$0.04} & 0.08{\scriptsize$\pm$0.13} & 0.45{\scriptsize$\pm$0.09} \\
\rowcolor{gray!12} E2E-LLM$^\dagger$ & 0{\scriptsize$\pm$0.00} & 0.6{\scriptsize$\pm$0.19} & 0.44{\scriptsize$\pm$0.10} & 0.215{\scriptsize$\pm$0.18} & \textbf{0.6{\scriptsize$\pm$0.11}} \\
\this & \textbf{0.9014{\scriptsize$\pm$0.17}} & \textbf{0.7217{\scriptsize$\pm$0.20}} & 0.43{\scriptsize$\pm$0.07} & \textbf{0.29{\scriptsize$\pm$0.29}} & 0.57{\scriptsize$\pm$0.10} \\
\bottomrule
\end{tabular}
\\[0.5mm]
{\scriptsize $^\dagger$The end-to-end baseline emits no per-claim evidence citations, hence zero coverage by construction.}
\vspace{-4mm}
\end{table}

\begin{figure}[!t]
    \centering
    \includegraphics[width=0.86\textwidth]{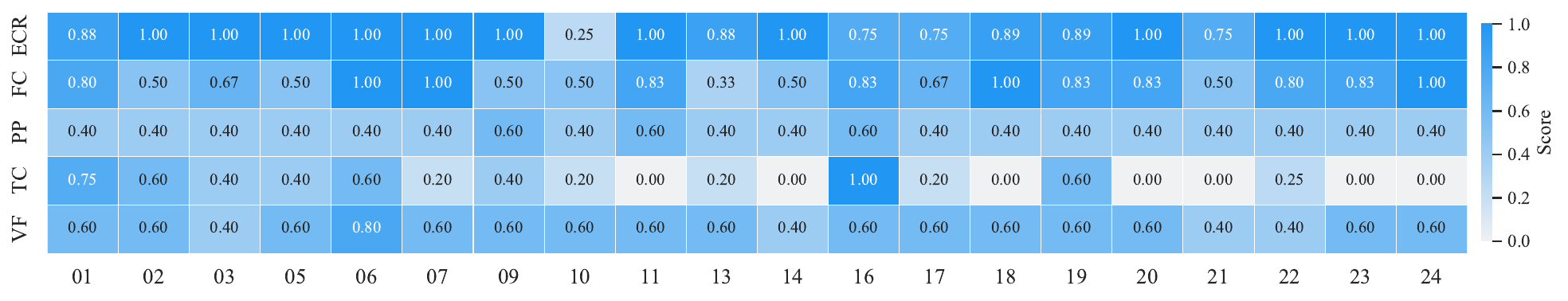}
    \caption{Per-scene scores of \this across five metrics.}
    \label{graph:heatmap}
    \vspace{-4mm}
\end{figure}

Table~\ref{table:main} reports means and standard deviations over the 20 main scenes; Figure~\ref{graph:heatmap} exposes the per-scene scores behind them.
The central result is the gap over the non-agentic baselines: the naive world-model family collapses on temporal coherence (TC 0.00--0.09---their clips rarely depict events in the evidentiary order) and reaches at most 0.6417 FC, with the single-modality variants confirming both modalities matter (dropping testimonies more than halves FC to 0.2983; dropping photographs costs grounding and order alike), and E2E-LLM produces no evidence citations at all (ECR 0.000), whereas \this grounds its timeline in 90.14\% of the evidence files on average and improves FC by 20.3\% (0.7217 vs.~0.6000) and TC by 34.9\% (0.2900 vs.~0.2150) over E2E-LLM---the traceability that matters most in a legal setting.

Removing either module individually lowers the scores (e.g., ECR drops from 0.9014 to 0.8436 without the fusion audit); PP and VF hover in a narrow band for all agent variants, as these dimensions are bound by the renderer rather than the orchestration~\cite{ICLR25-VideoPhy}.

\noindent\textbf{Process Evidence.}
Both agentic components stay consistently active: the probe flagged physical violations in all 18 probed scenes (1.61 corrective constraints injected on average), and the fusion audit ran the full 3 rounds in 19 of 20 scenes.

\begin{figure}[!t]
    \centering
    \includegraphics[width=0.99\textwidth]{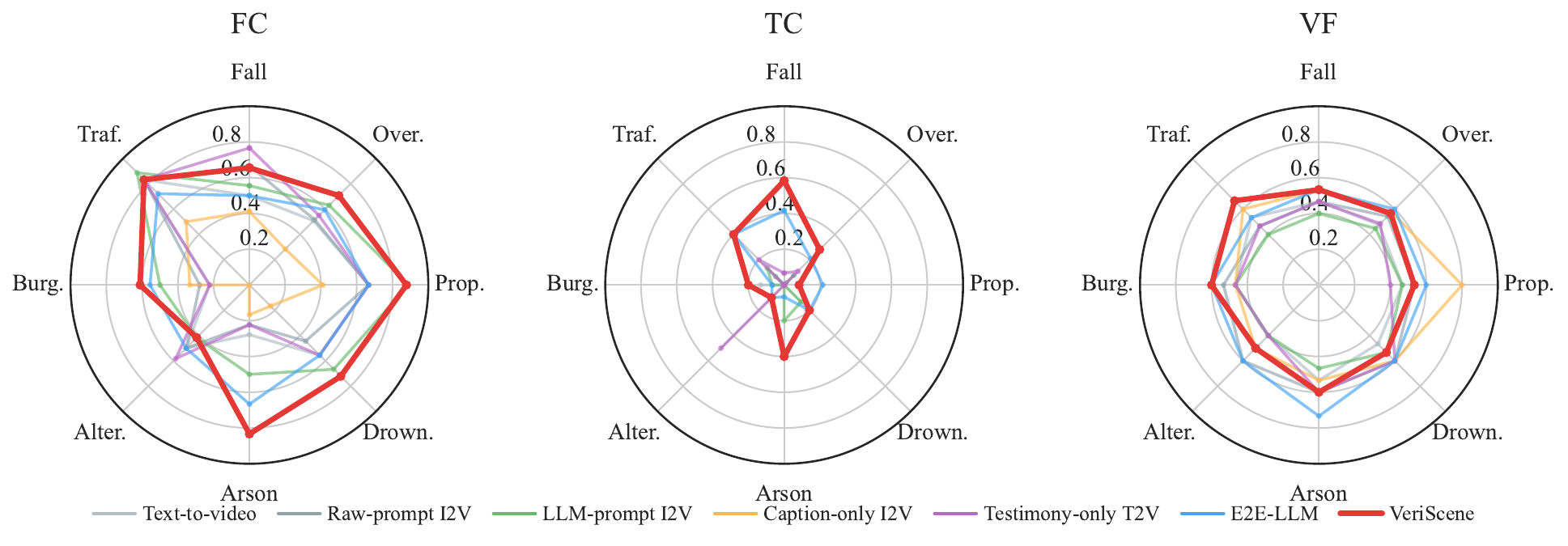}
    \caption{Per-metric comparison of \this and baselines by case type.}
    \label{graph:radar}
    \vspace{-4mm}
\end{figure}

Figure~\ref{graph:radar} breaks FC, TC, and VF down by case type---\this's envelope dominates every baseline on FC and TC, where the naive world-model family collapses toward the center---and Figure~\ref{graph:heatmap} details every scene: fall and arson scenes score highest, as their dynamics (gravity, flame propagation) are well captured by the probe and faithfully rendered, while water-related scenes are hardest---fluid interactions stress both the physical reasoning and the renderer, depressing PP and TC simultaneously.

\subsection{Qualitative Analysis} \label{section:evaluation:qualitative}
\begin{figure}[!t]
    \centering
    \begin{subfigure}{0.99\textwidth}
        \centering
        \includegraphics[width=\textwidth]{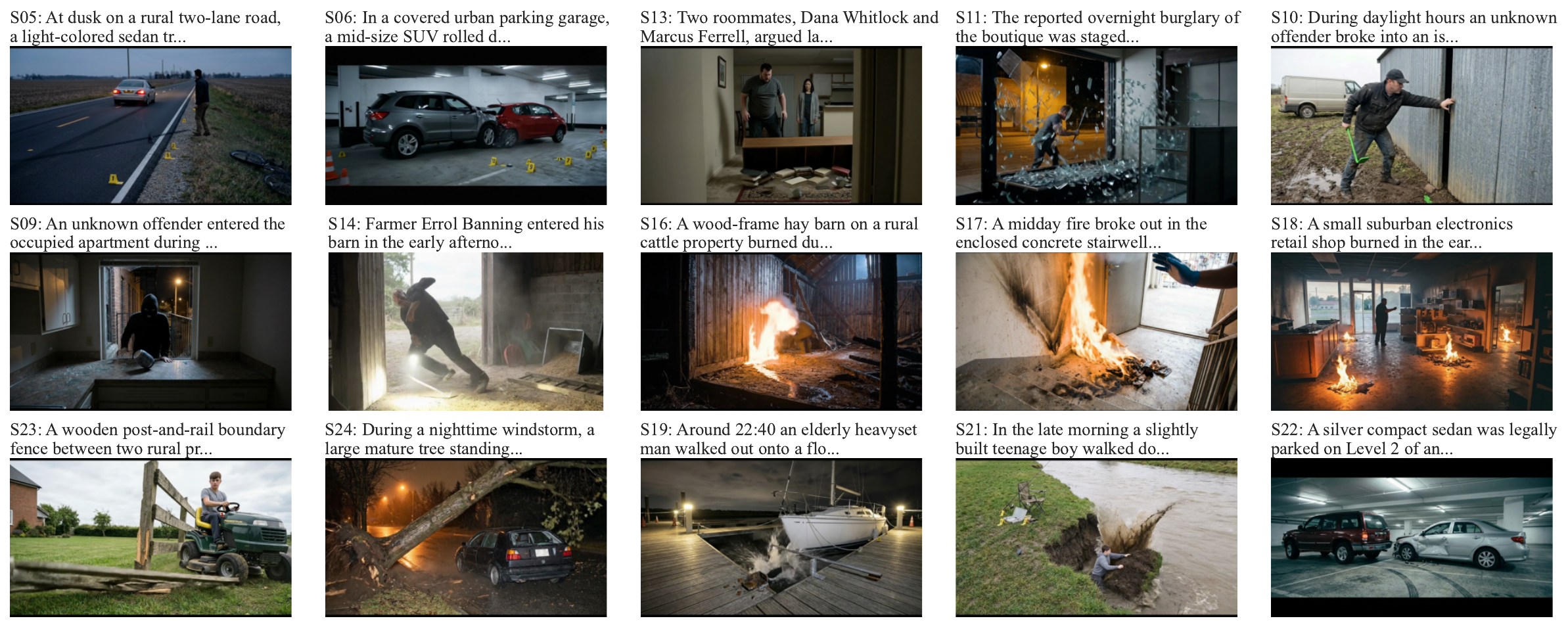}
        \caption{Re-enactment key frames of 15 representative scenarios.}
        \label{graph:qualitative}
    \end{subfigure}\\[0mm]
    \begin{subfigure}{0.99\textwidth}
        \centering
        \includegraphics[width=\textwidth]{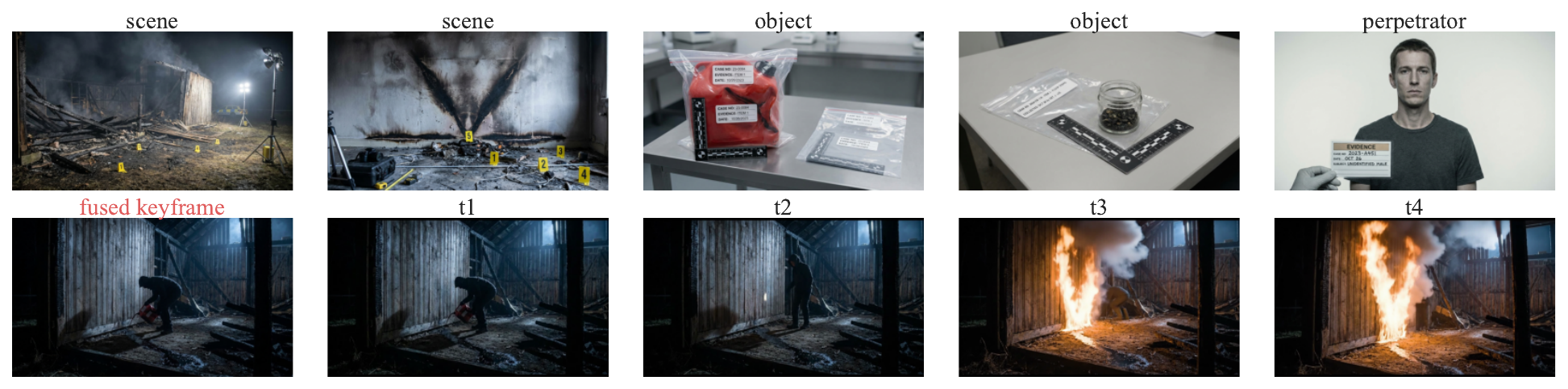}
        \caption{End-to-end reconstruction of an arson scenario.}
        \label{graph:casestudy}
    \end{subfigure}
    \caption{Qualitative results of \this.}
    \label{graph:qualstudy}
    \vspace{-4mm}
\end{figure}

\begin{figure}[!t]
    \centering
    \includegraphics[width=0.99\textwidth]{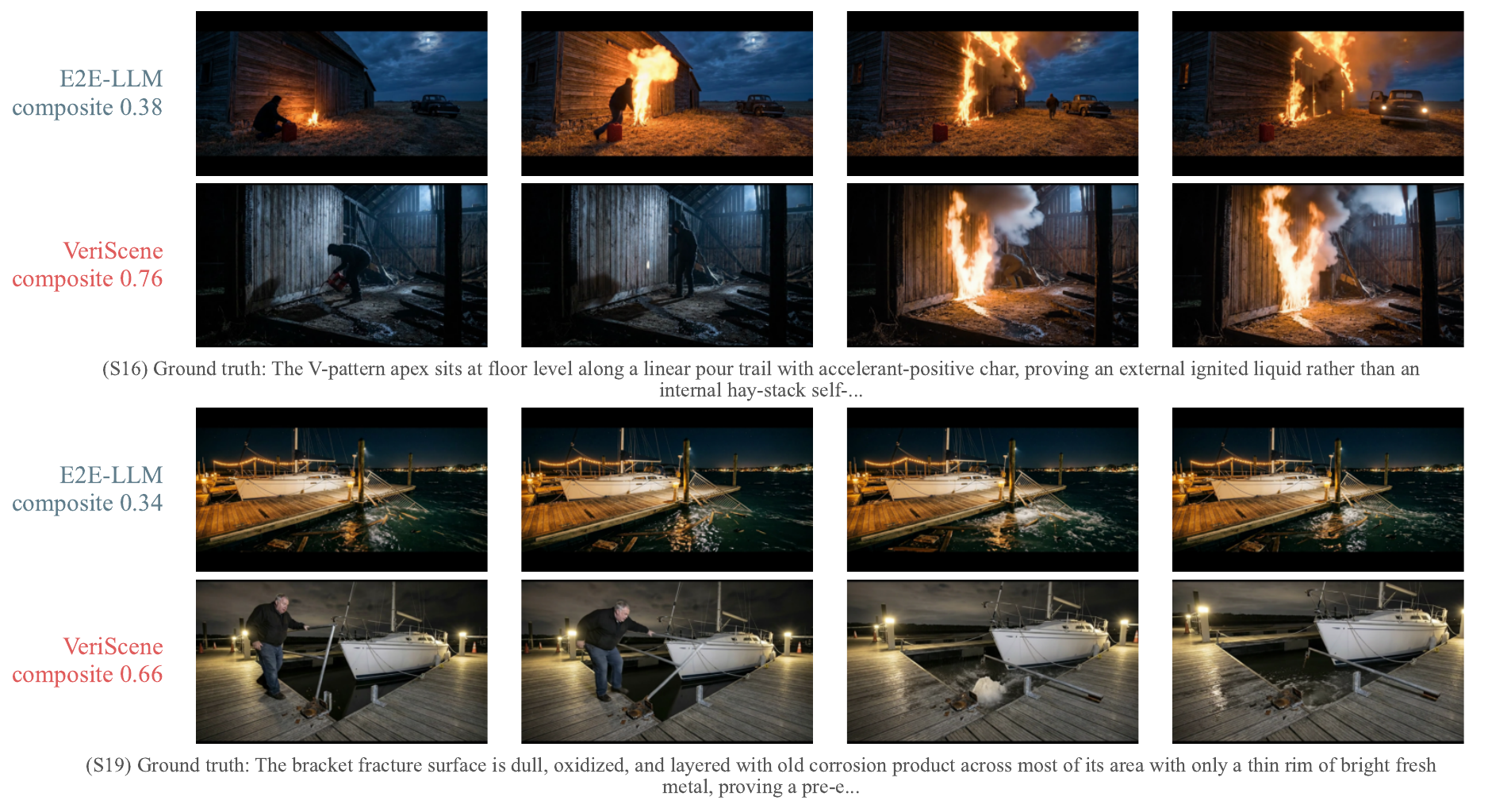}
    \caption{Re-enactments of E2E-LLM and \this on two scenarios.}
    \label{graph:compare}
    \vspace{-4mm}
\end{figure}

Figure~\ref{graph:compare} contrasts both pipelines on two scenarios: in the arson case E2E-LLM stages a generic exterior blaze untethered to the exhibits, whereas \this re-creates the evidenced indoor accelerant-trail fire; in the dock case E2E-LLM renders a static, unpopulated harbor while \this stages the documented railing failure and fall---in both, the agent keeps the rendered composite anchored to the ground-truth physical fact.

Figure~\ref{graph:qualitative} presents one key frame for 15 representative scenarios, showing coherent scene composition across case types from indoor falls to vehicle collisions.
Figure~\ref{graph:casestudy} walks through the S16 arson case end to end: from a bagged accelerant container and a V-shaped scorch photograph, \this renders the causal chain---pour, ignition, flame racing along the trail, fire climbing into the evidentiary V-pattern---turning disconnected exhibits into a physically ordered narrative a fact-finder can inspect frame by frame.

\subsection{Impact of the Orchestration LLM} \label{section:evaluation:llm}
\begin{table}[!t]
\centering
\caption{Quality, cost, and latency across orchestration LLM tiers.}
\label{table:llm}
\scriptsize
\setlength\tabcolsep{1.0mm}
\begin{tabular}{lccccccc}
\toprule
Brain & ECR & FC & PP & TC & VF & \$/scene & min \\
\midrule
Claude Opus & 0.9 & 0.6633 & 0.44 & \textbf{0.375} & \textbf{0.6} & 1.82 & 11.0 \\
\rowcolor{gray!12} Claude Sonnet & 0.8121 & 0.6467 & 0.42 & 0.245 & 0.54 & 1.65 & 8.1 \\
Gemini Pro & 0.8889 & 0.65 & \textbf{0.46} & 0.305 & \textbf{0.6} & 1.28 & 4.8 \\
\rowcolor{gray!12} Gemini Flash & \textbf{0.9764} & 0.6667 & 0.42 & 0.22 & 0.54 & 1.39 & 4.9 \\
Claude Haiku (weak) & 0.9361 & \textbf{0.6983} & 0.4 & 0.1175 & 0.48 & 0.13 & 5.8 \\
\bottomrule
\end{tabular}
\vspace{-4mm}
\end{table}

\begin{figure}[!t]
    \centering
    \begin{subfigure}{0.99\textwidth}
        \centering
        \includegraphics[width=\textwidth]{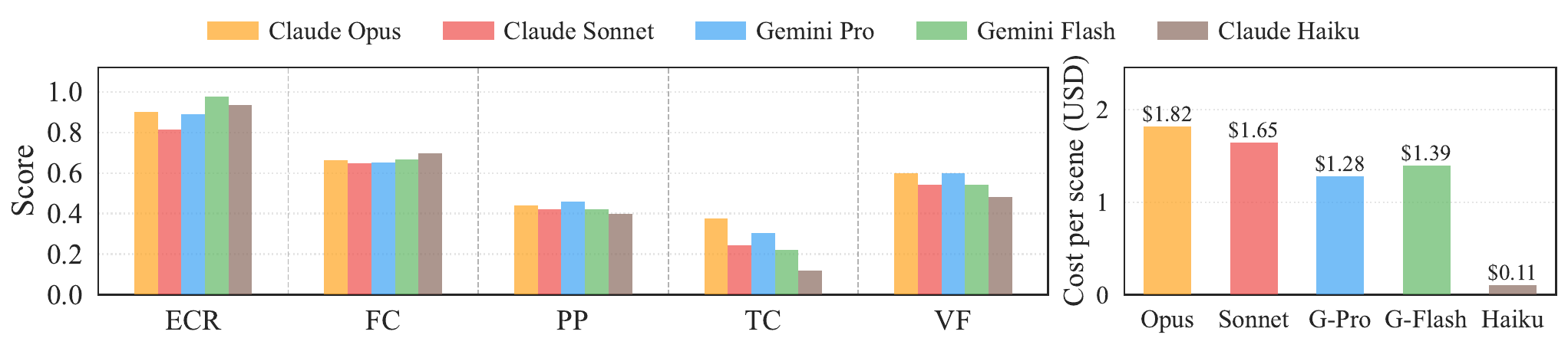}
        \caption{Quality and cost across orchestration LLMs.}
        \label{graph:llm}
    \end{subfigure}\\[0mm]
    \begin{subfigure}{0.99\textwidth}
        \centering
        \includegraphics[width=\textwidth]{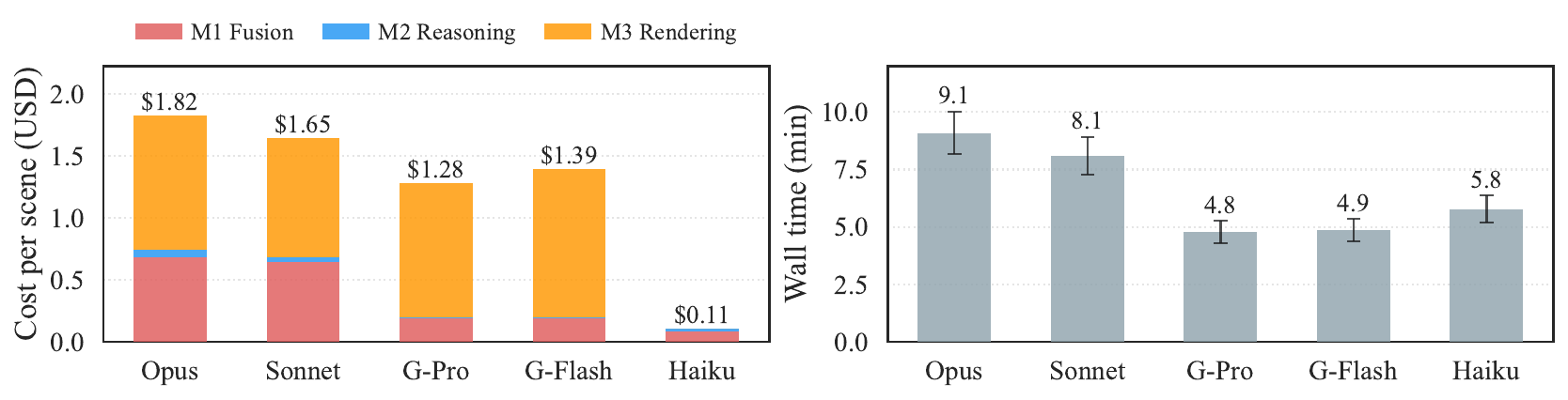}
        \caption{Per-module cost and wall time.}
        \label{graph:cost}
    \end{subfigure}
    \caption{Orchestration-LLM study on the 10-scene subset.}
    \label{graph:llmstudy}
    \vspace{-4mm}
\end{figure}

Finally, we swap the orchestration brain across five tiers (Table~\ref{table:llm}, Figure~\ref{graph:llm}).
Across the four commercial tiers, overall quality is remarkably stable (ECR 0.812--0.976, FC 0.647--0.667, PP/VF within 0.04/0.06)---suggesting the programmatic checks do the heavy lifting and insulate quality from the choice of brain.
The dimension most sensitive to brain capability is temporal coherence: Opus attains TC 0.375 versus 0.220 for Flash, so stronger reasoning primarily improves event ordering.
The weak tier pushes this to the extreme, pairing the cheapest brain (Claude Haiku) with a fully local renderer at \$0.13 per scene---14$\times$ cheaper than the Opus tier's \$1.82: the audit mechanism still carries it to 0.936 evidence coverage, but temporal coherence collapses to 0.118 and visual fidelity to 0.480---the checks preserve evidence grounding under weak brains, while temporal quality requires a stronger orchestrator.
Part of the weak tier's VF/PP drop is renderer-bound (it renders via the local world model).
Figure~\ref{graph:cost} decomposes cost and latency by module: with Opus, \this averages \$1.82 per scene at 9.1 minutes, dominated by rendering, while the flash tier is far cheaper at comparable quality---practical for high-volume triage.

\section{Discussion and Related Work}\label{section:discussion}

\noindent\textbf{Extensions: Sketch Maps, 3D Scans, and Audio Evidence.}
\revise{Investigative practice routinely produces a sketch map of the scene---where witnesses stood, where items were found, a vehicle's path---and increasingly 3D scene scans; both are natural added modalities for \this, serving as spatial priors that anchor evidence positions during fusion and keyframe synthesis.}
Multimodal LLMs likewise accept audio~\cite{Web-Gemini}, spoken statements joining the representation with semantic communication~\cite{BigCom24-SemCom} easing field transmission; the contradiction ledger exports resolved conflicts as cross-examination points.

\noindent\textbf{Limitations and Forged Evidence.}
The renderer imperfectly obeys text-level physics constraints~\cite{ICLR25-VideoPhy}; LLM-as-judge scoring carries biases~\cite{NeurIPS23-MTBench}, mitigated by rubrics; the synthetic benchmark (ethics precludes real case files) leaves transfer unverified; and \this models single-event scenes.
Generative models can also fabricate evidence, so deployment needs authenticity screening~\cite{CCS23-DEFAKE,TDSC26-PD2Net,TDSC26-ProactiveTrace}; \this's physical-consistency checks flag evidence whose dynamics violate world-model constraints.

\noindent\textbf{Legal Status and Ethical Considerations.}
\revise{A rendered re-enactment fits the category of \emph{demonstrative evidence}: under provisions such as section~68A of Singapore's Evidence Act~\cite{Web-EvidenceActSG}, it is admissible where it aids the court's comprehension of the primary evidence---contextualising the photographs and statements it is built from, never substituting for them; \this's per-claim citations provide the linkage such a tender requires.}
Persuasive-but-wrong renderings could bias fact-finders, so courtroom use still requires AI-generation disclosure, provenance labels~\cite{TCSVT26-AdaptiveRDH,TDSC26-ProactiveTrace}, and presentation alongside the primary evidence.

\noindent\textbf{Video Generation and World Models.}
Diffusion-based video generators~\cite{CVPR23-VideoLDM,Web-Veo} are increasingly framed as world simulators~\cite{Web-Cosmos,ICML24-Genie}: trained on Internet-scale video, they internalize approximate scene dynamics and serve as planning substrates for embodied agents.
Their objective, however, rewards visual plausibility rather than fidelity to any particular factual record, and physical-commonsense benchmarks expose systematic violations~\cite{ICLR25-VideoPhy}; \this therefore treats the world model as a renderer to be constrained---injecting evidence-derived physics rather than trusting unconstrained rollouts.

\noindent\textbf{Multimodal LLM Agents.}
Agent frameworks pair vision-language models~\cite{NeurIPS23-LLaVA} with tool use and self-refinement~\cite{NeurIPS23-Reflexion}; \this specializes them for evidence fusion.

\noindent\textbf{AI for Legal and Forensic Applications.}
Legal NLP focuses on textual tasks such as judgment prediction and benchmark reasoning~\cite{NeurIPS23-LegalBench,ACL20-LegalAI}; \this is, to our knowledge, the first to render incident scenes as videos from legal evidence.

\noindent\textbf{LLM-as-Judge and Physical Plausibility.}
Strong LLM evaluators track human preference despite biases~\cite{NeurIPS23-MTBench}; VideoPhy~\cite{ICLR25-VideoPhy} and PhyGenBench~\cite{ICML25-PhyGenBench} show generators often violate physical commonsense, whereas our benchmark demands consistency with a factual record.

\vspace{-2mm}
\section{Conclusion}\label{section:conclusion}
This paper presents \this, an LLM-agent pipeline that fuses forensic photographs and witness statements into a structured scene representation, derives world-model dynamics constraints, and renders the event as a re-enactment video, outperforming an end-to-end multimodal-LLM baseline on evidence coverage, factual consistency, and temporal coherence; we hope it establishes \revise{a demonstrative aid to the primary evidence---never a substitute for it}.

\subsubsection*{Acknowledgements.}
This research is supported by the National Research Foundation, Singapore and Infocomm Media Development Authority under its Trust Tech Funding Initiative. Any opinions, findings and conclusions or recommendations expressed in this material are those of the author(s) and do not reflect the views of National Research Foundation, Singapore and Infocomm Media Development Authority.

\vspace{-4.5mm}
\begingroup
\renewcommand{\small}{\fontsize{6.3}{7.1}\selectfont}
\bibliographystyle{splncs04}
\bibliography{input}
\endgroup

\end{document}